\documentclass[conference]{IEEEtran}
\IEEEoverridecommandlockouts
\usepackage{cite}
\usepackage{amsmath,amssymb,amsfonts}
\usepackage{algorithm}
\usepackage{algpseudocode}
\usepackage{graphicx}
\usepackage{textcomp}
\usepackage{setspace}
\usepackage{xcolor}
\usepackage{graphicx}
\usepackage{multicol}
\def\BibTeX{{\rm B\kern-.05em{\sc i\kern-.025em b}\kern-.08em
    T\kern-.1667em\lower.7ex\hbox{E}\kern-.125emX}}
\usepackage{caption}
\begin{document}
\bstctlcite{IEEEexample:BSTcontrol}
\title{Automatic hyperparameter selection in Autodock
{\footnotesize }
\thanks{}}

\author{\IEEEauthorblockN{Hojjat Rakhshani\textsuperscript{1}, Lhassane Idoumghar\textsuperscript{1}, Julien Lepagnot\textsuperscript{1}, Mathieu Br\'evilliers\textsuperscript{1}, Edward Keedwell\textsuperscript{2}}
	\IEEEauthorblockA{\textsuperscript{1}\textit{Universit\'e de Haute-Alsace, IRIMAS-UHA, F-68093 Mulhouse, France}
	\\
	\textsuperscript{2}\textit{University of Exeter, College of Engineering, Mathematics and Physical Sciences, Exeter EX4 4QF, UK}
	\\
		\textsuperscript{1}{\{hojjat.rakhshani, lhassane.idoumghar, julien.lepagnot, mathieu.brevilliers\}}@uha.fr
		\\
			\textsuperscript{2}{e.c.keedwell@exeter.ac.uk}
		}
\thanks{*This work is accepted  at 2018 IEEE International Conference on Bioinformatics and Biomedicine.}
}

\maketitle

\begin{abstract}
Autodock is a widely used molecular modeling tool which predicts how  small molecules bind to a receptor of known 3D structure. The current version of AutoDock uses meta-heuristic algorithms in combination with local search methods for doing the conformation search. Appropriate settings of hyperparameters in these algorithms are important, particularly for novice users who often find it hard to identify the best configuration. In this work, we design a surrogate based multi-objective algorithm to help such users by automatically tuning hyperparameter settings. The proposed method iteratively uses a radial basis function model and non-dominated sorting to evaluate the sampled configurations during the search phase. Our experimental results using Autodock show that the introduced component is  practical  and effective.
\end{abstract}

\begin{IEEEkeywords}
Surrogate-assisted
computation, Autodock, docking, multi-objective optimization
\end{IEEEkeywords}

\section{Introduction}

AutoDock is a protein-ligand docking tool which has now been distributed to more than 29000 users around the world\cite{trott2010autodock}. The application for this software arises from problems in computer-aided drug design. The ideal procedure would optimize the interaction energy between the substrate and the target protein.  It employs a set of meta-heuristic and local search optimization methods to meet this demand. The Autodock puts protein-ligand docking at the disposal of non-expert users how often do not know how to choose among the  dozens  of configurations for such algorithms. This motivated us to bring  the  benefits of hyperparameter tuning to users of AutoDock.

\par
The global optimization problem arises in many real-world applications as well as molecular docking. A standard continuous optimization problem seeks a parameter vector \( \mathbf x^{*}\) to minimize an objective function \(f(\mathbf x): \mathbb{R}^{D}\rightarrow \mathbb{R}\), i.e. \(f(\mathbf x^{*}) < f(\mathbf x)\) for all \(\mathbf x \in \Omega\), where \(\Omega=\mathbb{R}^{D}\) denotes the search domain (a maximization problem can be obtained by negating the \(f)\).  Over the years, optimization algorithms have been effectively applied to tackle this problem. These algorithms, however, may need to be fine-tuned which is a major challenge to their successful application. This is primarily due to the highly nonlinear and complex properties associated with the objective function. In this context, the advantages of algorithm configuration techniques become clear. 
\par
The hyperparameter tuning domain has been dominated by model-based optimization \cite{hutter2011sequential} that adopt probabilistic surrogate models to replace in part the original computationally expensive evaluations. They construct computationally cheap to evaluate surrogate models in order to provide a fast approximation of the expensive fitness evaluations during the search process.
Surrogate based techniques try to model conditional probability \(p(y|\varphi)\) of a \(m-\)dimensional configuration \(\varphi\) given \(n\) observations \(\textbf{S}\):
\begin{equation}
\resizebox{.90\hsize}{!}{$\mathbf{S}=\left [ \mathbf{x}^{(1)},...,\mathbf{x}^{(n)} \right ]^{\textup{T}} \in \mathbb{R}^{n\times m}, \mathbf{x}=\left \{ x_{1},...,x_{m} \right \} \in \mathbb{R}^{m}$}
\label{eq_2}
\end{equation}
with the corresponding evaluation metrics \(\mathbf{y}\):

\begin{equation}
\resizebox{.90\hsize}{!}{$\mathbf{y}=\left [y^{(1)},...,y^{(n)} \right ]^{\textup{T}}=\left [y(\mathbf{x}^{(1)}),...,y(\mathbf{x}^{(n)}) \right ]^{\textup{T}} \in \mathbb{R}^{n}$}
\label{eq_3}
\end{equation}

\par
The essential question arises in model based algorithms is which individuals should be chosen to be evaluated using the exact fitness function. This is most characterized by making a good balance between the exploration and exploitation capabilities. One of the earliest researches in this direction was performed by Jones et al. who adopted a Kriging surrogate to fit the data and make a balance between exploration and exploitation \cite{jones1998efficient}. The exploration property of their proposed efficient global optimization (EGO) algorithm is enhanced by the fact that the expected improvement (EI) (an acquisition function) is conditioned on points with large uncertainty and low values of the surrogate. In another study, Booker et al. \cite{booker1999rigorous} sought out a balanced search strategy by taking into account sample points with low surrogate predictions and high mean square error. Moreover, Wang et al. \cite{wang2004mode} introduced the mode-pursuing approach which favors the trial points with low surrogate values according to a probability function. Regis and Shoemaker \cite{regis2005constrained} also put forward an approach according to which the next candidate point is chosen to be the one that minimizes the surrogate value subject to distance from previously evaluated points. The distance starts from a high value (global search) and ends with a low value (local search). They also proposed Stochastic Response Surface (SRS) \cite{regis2007stochastic} algorithm that cycles from emphasis on the objective to emphasis on the distance using a weighting strategy. The SRS, however, mitigated some of the requirements for inner acquisition function optimization. For this reason, we focused on proposing a new algorithm configuration approach based on SRS model. The proposed algorithm, called as MO-SRS, modify the SRS so as to be able to handle multi-objective problems. More precisely, MO-SRS is equipped with the idea of multi-objective particle swarm optimization (PSO)\cite{1004388}. We used the MO-SRS to make a balance between intermolecular energy and the Root Mean Square Deviation (RMSD) during the hyperparameter optimization in Autodock.

\par
The rest of the paper is organized as follows. Section II provides us with a brief review on the related works. Section III gives a brief description for the docking problem. Section IV elaborates technical details of our proposed approach.  In Section V, the performance of the introduced components is investigated by conducting a set of experiments. The last section summarizes the paper and draws conclusions.

\section{Related works}

Sequential Model-based Algorithm Configuration (SMAC) \cite{hutter2011sequential}, Hyperband \cite{li2017hyperband}, Spearmint \cite{snoek2012practical}, F-race \cite{birattari2010f} and Tree-structure Parzen Estimator (TPE) \cite{bergstra2011algorithms} are examples of well known methods for hyperparameter optimization. SMAC adopted a random forests model and Expected Improvement (EI) to compute \(p(y|\varphi)\). Similarly, TPE et al. defined a configuration algorithm based on tree-structure Parzen estimator and EI. To tackle the \textit{curse of dimensionality}, TPE assigns particular values of other elements to the hyperparameters which are known to be irrelevant. Ilievski et al. \cite{ilievski2017efficient} proposed a deterministic method which employs dynamic coordinate search and radial basis functions (RBFs) to find most  promising configurations. By using the RBFs  \cite{park1991universal} as surrogate model, they mitigated some of the requirements for inner acquisition function optimization. In another work \cite{snoek2015scalable}, the authors put forward  neural networks  as  an  alternative  to  Gaussian process  for  modeling  distributions  over  functions. Interestingly, \textit{Google} introduced \textit{Google Vizier} \cite{golovin2017google}, an internal service for surrogate based optimization which incorporates Batched Gaussian Process Bandits along with the EI acquisition function. 
\par
Although the above-mentioned approaches have
been proven successful, they are not able to handle several objective functions during the tuning process. Consequently, this paper presents a novel multi-objective algorithm  which integrates the idea of a multi-objective meta-heuristics with the SRS method to find a subset of promising configurations.
\section{Molecular docking}
Molecular docking enables us to find an optimal conforma-
tion between a ligand $ L $ and receptor $ R $. In another word, it predicts the preferred orientation of $ L $ to R when bound to each other in order to form a stable complex. A schematic example of this procedure is illustrated in Fig.1. This problem can be formulated as a multi-objective problem consisting of minimizing the RMSD score and the binding energy $ E_{inter} $. In Autodock, energy
function $ E_{inter} $ is defined as follows \cite{trott2010autodock}: 

\begin{figure}[htbp]
	\centerline{\includegraphics[width=0.35\textwidth]{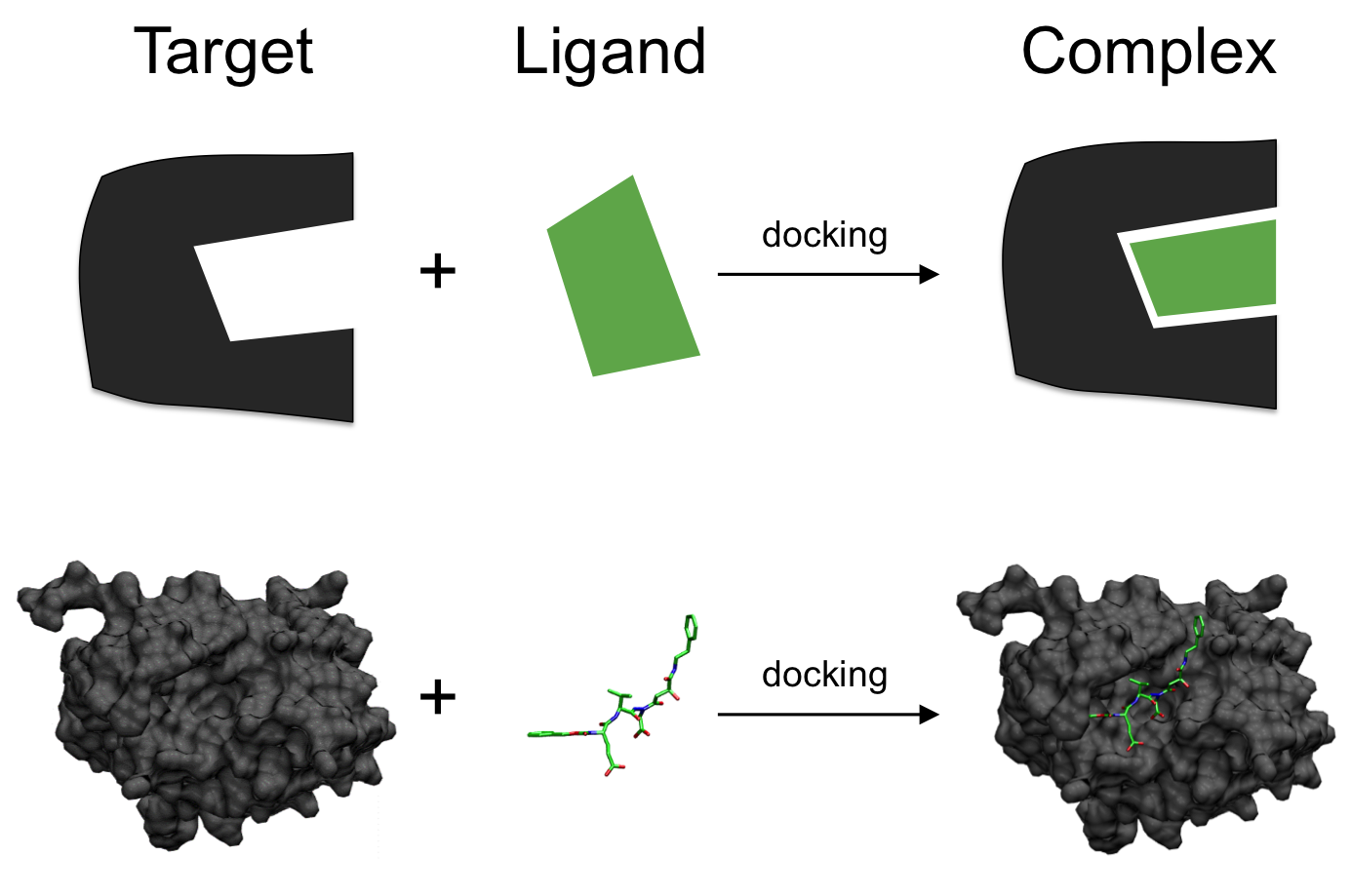}}
	\caption{Schematic example of docking a ligand to a protein}
		\vspace{-0.5cm}
\end{figure}

\begin{equation}
E_{inter}=Q_{bound}^{R-L}+Q_{unbound}^{R-L}
\end{equation}

\begin{multline*}
Q=W_{vdw}\sum_{i,j} \left ( \frac{A_{ij}}{r_{ij}^{12}}-\frac{B_{ij}}{r_{ij}^{6}}  \right )+
\\
W_{hbond}\sum_{i,j} E(t) \left ( \frac{C_{ij}}{r_{ij}^{12}}-\frac{D_{ij}}{r_{ij}^{10}}  \right )+
\\
W_{elec}\sum_{i,j} \frac{q_{i}q_{j}}{\varepsilon (r_{ij})r_{ij})}+W_{sol}\sum_{i,j}(S_{j}V_{i}+S_{i}V_{j})e^{(-\frac{r_{ij}^{2}}{2\sigma^{2} })}
\end{multline*}

In (3),
$ Q_{bound}^{R-L} $ and $ Q_{unbound}^{R-L} $ represent the states of ligand-protein complex in bound and unbound modes, respectively.

The  pairwise  energetic  terms  take into the account  evaluations for repulsion ($ vdw $), electrostatics ($ elec $), desolvation ($ sol $) and hydrogen bonding ($ hbond $). The constant weights $ W_{vdw} $, $ W_{elec} $, $ W_{hbond} $ and $ W_{sol} $ belong to Van der Waals, electrostatic interactions, hydrogen bonds and desolvation, respectively. Moreover, $ r_{ij} $, $ A_{ij} $, $ B_{ij} $, $ C_{ij} $ and $ D_{ij}$ denotes Lennard-Jones parameters, while $ E(t) $ function is the angle-dependent directionality. Also, $ V $ is the volume  of  atoms that  surround  a  given  atom, weighted by a solvation parameter ($ S $). For a detailed report of all the variables please refer to \cite{morris2009autodock4}.

\section{The proposed method}
This section discusses in detail the main components of the proposed MO-SRS. We extend the idea of stochastic RBF to the multi-objective case to be suitable for the Autodock application. Compared to the evolutionary algorithms like genetic algorithm, MO-SRS need less computational time by virtue of surrogate modeling techniques. A workflow of the introduced approach is illustrated in Fig 2.

\begin{figure}[htbp]
	\centerline{\includegraphics[width=0.35\textwidth]{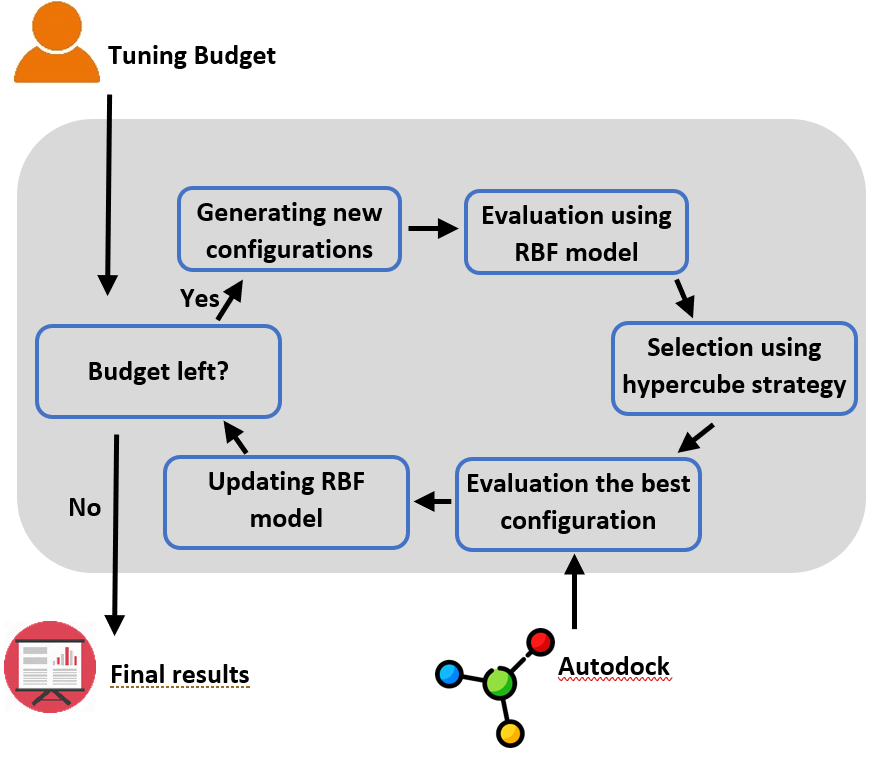}}
	\caption{Workflow of the introduced automatic hyperparameter tuning approach for Autodock}
	\vspace{-0.5cm}
\end{figure}

The MO-SRS starts the optimization procedure by generating a set of random configurations (i.e., initial population). During this phase, it might be possible to miss a considerable portion of the promising area due to the high dimensionality of the configuration space. It should be noticed that we have a small and a fix computational budget and increasing size of the initial population cannot remedy the issue. Furthermore, it is crucial for a model-based algorithm to efficiently explore the search space so as to approximate the nonlinear behavior
of the objective function. The Design of Computer Experiments (DoCE) methods are often used to partially mitigate high dimensionality of the search space. Among them, MO-SRS utilized the Latin Hypercube Sampling (LHS) \cite{mckay1979comparison} which is a DoCE method for providing a uniform cover on the search space using a minimum number of population. The main advantage of LHS is that it does not require an increased initial population size for more dimensions.

As the next step, we evaluate all the generated configurations \(\mathbf{x}_{i}(i=1,2,\cdots,n)\) using the Autodock to yield outcomes \(y_{i}^{(1)}=f^{(1)}(\mathbf{x_{i}})\) and \(y_{i}^{(2)}=f^{(2)}(\mathbf{x_{i}})\). Here, $ {f_{1}} $ and $ {f_{2}} $ denote the energy and RMSD function variables, respectively. Thereafter, we adopt surrogate model techniques to approximate the fitness function using this data set. 
\par
Surrogate models offer a set of mathematical tools for predicting the output of an expensive objective function \(f\). Particularly, they try to predict the fitness function \(f\) for any unseen input vector \(\hat{x}\) according to computed data points \((x_{i},y_{i})\). Given solution \(\boldsymbol{\hat{x}}\) and objective function  \(f\), a surrogate model \(\hat{f}\) can be defined as in (4), where \(\epsilon \) is the approximation error. 

\begin{equation}
\tilde{f}(\boldsymbol{\hat{x}})=f(\boldsymbol{\hat{x}})+\epsilon  \label{eq}
\end{equation}

\par

Among different surrogate models, we utilized the Radial Basis Function (RBF) which is a good model for high dimensional problems \cite{diaz2011selection,park1991universal}. It belongs to an
interpolation method for scattered multivariate data which considers all the sample points. In this light, it introduces linear combinations based on a RBF \(h(\mathbf{x})\) to approximate the desired response function \(f\); as presented in (5).

\begin{equation}
\tilde{f}(\hat{\mathbf{x}})=\sum_{i=1}^{n} w_{i}  h(\left | \hat{\mathbf{x}}-\mathbf{x}_{i} \right |) + \mathbf{b}^{\top}\hat{\mathbf{x}}+a  \label{eq}
\end{equation}

In (5), \(w_{i}\) shows \(i^{th}\) unknown weight coefficient, \(h\) determines a radial basis function, \(\hat{\mathbf{x}}\) is an unseen point and \(\epsilon\) are independent errors with a variance \(\sigma^{2}\). A radial function \(h: \mathbb{R}^{m} \rightarrow \mathbb{R}\) has the property \(h(\mathbf{x})=h(\left \| x \right \|)\). By having a suitable kernel \(h\); the unknown
parameters \(a, \mathbf{b}\) and \(\mathbf{w}\) could be obtained by solving the
following system of linear equations:
\begin{equation}
\mathbf{}
\begin{pmatrix}
\mathbf{\Phi}  & \mathbf{P}\\ 
\mathbf{P}^{\top}& 0
\end{pmatrix} \begin{pmatrix}
\mathbf{w}
\\ 
\mathbf{c}
\end{pmatrix}=\begin{pmatrix}
f_{\psi}
\\ 
0
\end{pmatrix}
\label{eq}
\end{equation}

Here, \(\mathbf{\Phi}\) is a \( n \times n \) matrix with \(\Phi_{i,j}=h(\left \| \mathbf{x}_{i} - \boldsymbol{x}_{i} \right \|)\), \(\mathbf{c}=(\mathbf{b}^{\top}, a)^{\top}\) and

\begin{equation}
\mathbf{}
\mathbf{P}^{\top}=\begin{pmatrix}
\mathbf{x}_{1} & \mathbf{x}_{2} & \cdots  & \mathbf{x}_{n} \\ 
1 & 1  & \cdots  & 1
\end{pmatrix}
\label{eq}
\end{equation}
The linear system (6) has a unique solution and it can be used to approximate function values.

As opposite to the standard SRS method, we have two main objective functions which should be optimized: the intermolecular energy and the RMSD. Hence, we repeat the above mentioned procedure for both the objective functions. More precisely, we will train two surrogate models  $ \hat{f_{1}} $ and $ \hat{f_{2}} $ to approximate the energy and RMSD output variables, respectively.

Now we are ready to use the trained surrogate models inside a pre-selection strategy. Accordingly, we generate a set of $ N $ candidates to be evaluated using $ \hat{f_{1}} $ and $ \hat{f_{2}}$. In the single-objective SRS code, the following equation generates neighborhoods using the current best solution $\hat{f_{2}}$, a randomly produced vector $ \textup{{v}} \in \mathbb{R}^{m} $ and adaptive parameter $ \gamma $:

\begin{equation}
\boldsymbol{\sigma}_{j}=\boldsymbol{x}_{best}+\gamma \otimes \textup{{v}},	j=1... N\label{eq}
\end{equation}

Generally speaking, in multi-objective optimization the best solutions need to be chosen based on two objectives and there is no single best solution. So, we cannot directly apply the standard search operators of the SRS. Moreover, we have to update our surrogate models $ \hat{f_{1}} $ and $ \hat{f_{2}}$ based on a best obtained solution which is another problem for the MO-SRS. Having this in mind, we borrowed the idea of leader selection in multi-objective PSO \cite{1004388} algorithm which handles the same problem (update equation of the PSO also depends on the global best solution and finding a leader particle is an important task). Inferred from \cite{1004388}, MO-SRS first builds an archive which contains the \textit{Pareto-optimal} solutions.  A solution vector is \textit{Pareto-optimal} if there is not another solution that dominates it. In mathematical terms, solution $x_{1}$ is said to dominate another solution $x_{2}$, if:

\begin{itemize}
	
	\item$  f_{i}(x_{1}) \leqslant f_{i}(x_{2}) $ for all $ i \in \left \{ 1,2,...,k \right \} $ and
	
	\item $ f_{i}(x_{1}) \leqslant f_{i}(x_{2}) $ for at least one $  i \in \left \{ 1,2,...,k \right \} $
	
\end{itemize}

The MO-SRS divides the objective space into hypercubes and assigns a fitness value to each hypercube based on the number of Pareto-optimal solutions that lie in it. Then, it employs the roulette-wheel to find the superior hypercube.  A randomly selected solution from that hypercube is determined to be the best solution $ \boldsymbol{x}_{best} $.
All the $N$ candidates are evaluated by the previously trained surrogates $ \hat{f_{1}} $ and $ \hat{f_{2}} $. Again, we select the best solution according to the above mentioned strategy. This best solution is then used to update our surrogate models. 
All the aforementioned steps continue until some stopping criteria are met. To sum up, MO-SRS approach is  illustrated in Fig. 3. The MO-SRS will be available soon on https://github.com/ML-MHs/Auto-Autodock.

\begin{algorithm}[H]
    \centering \normalsize
	\caption{MO-SRS} \label{alg:MyAlgorithm}
	\begin{spacing}{0.8}
		\begin{algorithmic}[0.6]
			\State $\text{Use LHS to initialize a population } \boldsymbol{X} \leftarrow \left \{ \mathbf{x}_{1},\cdots,\mathbf{x}_{n_{0}} \right \} $
			\State $n \gets n_{0}$
			\State $y_{i}^{(1)}=f^{(1)}(\mathbf{x_{i}}): i\leftarrow 1\cdots n $
			\State $y_{i}^{(2)}=f^{(2)}(\mathbf{x_{i}}): i\leftarrow 1\cdots n $			
			\Repeat
			\State Find $ \textit{Pareto-optimal } P \text{ using } y_{i}^{(1)} \text{ and } y_{i}^{(2)} $			
			\State Set $ \boldsymbol{x}_{best} $ using $ P $ and described hypercube technique			
			\State Fit surrogate  $ \hat{f}^{(1)}$ by $ \mathcal{B}_{n} \leftarrow \left \{ \left ( x_{i},y_{i}^{(1)} \right )  :i\leftarrow 1\cdots n \right \} $
			\State Fit surrogate  $ \hat{f}^{(2)}$ by $ \mathcal{C}_{n} \leftarrow \left \{ \left ( x_{i},y_{i}^{(2)} \right )  :i\leftarrow 1\cdots n \right \} $			
			\State Generate new solutions $\boldsymbol{\sigma}_{j} : j\leftarrow 1\cdots N $ as in (8)
			\State Apply the bound constraints on each solution $\boldsymbol{\sigma}_{j}$ 
			\State $\hat{y}_{j}^{(1)}=\hat{f}^{(1)}(\mathbf{\sigma_{j}}): j\leftarrow 1\cdots N $
			\State $\hat{y}_{j}^{(2)}=\hat{f}^{(2)}(\mathbf{\sigma_{j}}): j\leftarrow 1\cdots N $
			\State Find $ \textit{Pareto-optimal } \hat{P} \text{ using } \hat{y}^{(1)} \text{ and } \hat{y}^{(2)} $
			\State Set $ \boldsymbol{x}_{n+1} $ using $ \hat{P} $ and described hypercube technique
			
			\State $y_{n+1}^{(1)}=f^{(1)}(\mathbf{x_{n+1}})$
			\State $y_{n+1}^{(2)}=f^{(2)}(\mathbf{x_{n+1}})$	
			\State $\mathcal{B}_{n+1}\leftarrow \mathcal{B}_{n} \cup \left \{ (\mathbf{x}_{n+1},y_{n+1}^{(1)}) \right \} $
			\State $\mathcal{C}_{n+1}\leftarrow \mathcal{C}_{n} \cup \left \{ (\mathbf{x}_{n+1},y_{n+1}^{(2)}) \right \} $
			\State $n\leftarrow n+1$
			\Until{stopping criteria are met}
		\end{algorithmic}
	\end{spacing}
\end{algorithm}
\begingroup
\centerline{Fig. 3. Pseudocode of the MO-SRS method}
\endgroup

\section{Experimental results}
In this section, the performance of MO-SRS hyperparameter tuning method on Autodock task is investigated. We considered two scenarios from Autodock documentations which should be minimized: 1)1DWD and 2) HSG1. All simulations are performed using Matlab. We adopted the same implementation and parameter configuration for the SRS as suggested in \cite{muller2014matsumoto}. The Cubic function is used as the kernel in the RBF.
\par
We applied the MO-SRS to tune the hyperparameters of Autodock for the genetic algorithm (GA) \cite{morris1998automated}, simulated annealing (SA) \cite{kirkpatrick1983optimization}, a local search (LS) called as pseudo-Solis-Wet \cite{trott2010autodock} and a hybrid method (HB) which combines the GA and LS. The GA is a well-known evolutionary optimization algorithm which is highly sensitive ot the initial value of its parameters. The performance of the GA is dependent on the structure of the problem at hand and do not scale well with complexity. Consequently, parameters of GA like crossover, population size and mutation should be chosen with care. A small population size will lead to the early convergence problem, while using a large population increases the computational cost. The configuration tuning of GA becomes even more challenging when there is a  correlation between its hyperparameters. For example, mutation is more effective on smaller population sizes, while the crossover is likely to benefit from large populations. All the mentioned reasons make the GA a suitable algorithm for benchmarking the performance of MO-SRS. The same situation could happen for the considered LS and SA algorithms. From other points of view, the adopted algorithms are introduced to measure search performance of the MO-SRS under different dimensions. The LS and GA belong to low dimensional problems, SA is a medium dimension and HB algorithm is a high dimensional hyperparameter tuning problem. A detailed information for the optimized configurations for GA, SA and LS are presented in Tables I-III, respectively.
\par
The obtained results are reported in Tables IV and V. The MO-SRS should obtain a reliable search performance using a limited computational budget and so the number of evaluations is set to 100. For each algorithm, the search bounds are reported in Tables I-III. To reduce the influence of stochastic error, experiments are repeated 10 times for each problem. In Tables IV-V, the \textit{tuned } prefix  denotes the optimized algorithms using the MO-SRS. The MO-SRS offers a set of final solutions for each of the cases and we used a \textit{Pareto-optimal} graph to illustrate such solutions; as depicted in Fig. 4. However, it should be noticed that the results offered by the default methods in Autodock are based on single-objective algorithms. For this reason, we compared the obtained results of the MO-SRS according to each of those objectives. The subindex 1 in tables IV-V shows the performance of the compared algorithms for the binding energy and subindex 2 denotes the same for the RMSD criterion. The results are averaged over 10 runs. The energy unit is in kcal/mol and RMSD unit is in Angstroms. 
\par
The results of the first case study are presented in Table IV. As it can be seen, MO-SRS converges closer to the global optimum. In the case of LS method, we can see that the tuned algorithm yields a minimized energy -17.1530 and RMSD 0.0860. In docking domain, RMSD $ < $ 1.5 Angstrom is always wise to consider. Similarly, MO-SRS provides more accurate results for the second scenario according to the results in Table V. 
\par
In these tables, we can see how the introduced method tries to find more than one objective simultaneously. 
In this regard, there are two main points. The first one is that the diversity of the obtained solutions strongly depends on the performance of the algorithm at hand. For example, LS performs better than the GA and consequently we can see the  \textit{Pareto-optimal} for LS contains more solution. The second point is that all \textit{Pareto-optimal} solutions are obtained  only after 100 evaluations. This confirms our hypothesis that MO-SRS can effectively be use to tune hyperparameters of the Autodock. The main components of MO-SRS are the adopted multi-objective and surrogate modeling techniques which can be further improved in the future works.

\begin{table}[H]
    \centering \small 
    \captionsetup{font=scriptsize}
	\caption{Details for the optimized hyperparameters of GA}
	\begin{center}	
		\scalebox{0.75}{	
	\begin{tabular}{|l|l|l|l|}
		\hline
		Hyperparameter      & Type       & Range            & Default \\ \hline
		seed                & Integer    & {[}0,10000000{]} & Random  \\ \hline
		ga\_pop\_size       & Integer    & {[}50,500{]}     & 150     \\ \hline
		ga\_elitism         & Binary     & 0,1              & 1       \\ \hline
		ga\_mutation\_rate  & Continuous & {[}0.2,0.99{]}   & 0.02    \\ \hline
		ga\_crossover\_rate & Continuous & {[}0.2,0.99{]}   & 0.80    \\ \hline
	\end{tabular}}
\end{center}
\end{table}

\begin{table}[H]
    \centering \small 
    \captionsetup{font=scriptsize}
	\caption{Details for the optimized hyperparameters of SA}
	\vspace{-0.2cm}
	\begin{center}	
		\scalebox{0.70}{	
	\begin{tabular}{|l|l|l|l|}
		\hline
		Hyperparameter   & Type       & Range             & Default \\ \hline
		seed             & Integer    & {[}0,10000000{]}  & Random  \\ \hline
		tstep            & Continuous & {[}-2.0,2.0{]}    & 2.0     \\ \hline
		qstep            & Continuous & {[}-5.0,5.0{]}    & 2.0     \\ \hline
		dstep            & Continuous & {[}-5.0,5.0{]}    & 2.0     \\ \hline
		rtrf             & Continuous & {[}0.0001,0.99{]} & 0.80    \\ \hline
		trnrf            & Continuous & {[}0.0001,0.99{]} & 1.0     \\ \hline
		quarf            & Continuous & {[}0.0001,0.99{]} & 1.0     \\ \hline
		dihrf            & Continuous & {[}0.0001,0.99{]} & 1.0     \\ \hline
		accs             & Integer    & {[}100,30000{]}   & 30000   \\ \hline
		rejs             & Integer    & {[}100,30000{]}   & 30000   \\ \hline
		linear\_schedule & Binary     & 0,1               & 1       \\ \hline
	\end{tabular}}
\end{center}
\end{table}

\begin{table}[H]
    \centering \small 
    \captionsetup{font=scriptsize}
	\caption{Details for the optimized hyperparameters of the adopted local search}	
	\vspace{-0.2cm}
	\begin{center}	
		\scalebox{0.6}{
	\begin{tabular}{|l|l|l|l|}
		\hline
		Hyperparameter & Type    & Range            & Default \\ \hline
		seed           & Integer & {[}0,10000000{]} & Random  \\ \hline
		sw\_max\_its   & Integer & {[}100,1000{]}   & 300     \\ \hline
		sw\_max\_succ  & Integer & {[}2,10{]}       & 4       \\ \hline
		sw\_max\_fail  & Integer & {[}2,10{]}       & 4       \\ \hline
	\end{tabular}}
\end{center}
\end{table}

\vspace{-0.7cm}

\begin{table}[H]
    \centering \small 
    \captionsetup{font=scriptsize}
	\caption{The average docking results of the SA, GA, LS and HB algorithms after 10 runs for the first case study.}	
	\vspace{-0.2cm}
		\begin{center}	
			\scalebox{0.6}{	
	\begin{tabular}{|l|l|l|}
		\hline
		Algorithm \hspace{1cm} & Energy \hspace{1cm}  & RMSD \hspace{1cm}  \\ \hline
		Tuned $ \text{SA}^{*}_{1} $      & -13.4990 & 4.0450 \\
		Tuned $ \text{SA}^{*}_{2} $      & -10.9280 & 3.4140 \\
		$ \text{SA}_{1} $       & -10.9260 & 3.5460 \\
		$ \text{SA}_{2} $       & -10.9260 & 3.5460 \\ \hline
		Tuned $ \text{GA}^{*}_{1} $      & -13.7110 & 3.4960 \\
		Tuned $ \text{GA}^{*}_{2} $      & -11.9070 & 2.4550 \\
		$ \text{GA}_{1} $       & -10.3610 & 5.2730 \\
		$ \text{GA}_{2} $       & -10.1120 & 3.2880 \\ \hline
		Tuned $ \text{LS}^{*}_{1} $      & -17.4270 & 0.2190 \\
		Tuned $ \text{LS}^{*}_{2} $      & -17.1530 & 0.0860 \\
		$ \text{LS}_{1} $       & -17.2990 & 0.1980 \\
		$ \text{LS}_{2} $       & -17.1920 & 0.1400 \\ \hline
		Tuned $ \text{HB}^{*}_{1} $      & -13.7110 & 3.4960 \\
		Tuned $ \text{HB}^{*}_{2} $      & -11.9070 & 2.4550 \\
		$ \text{HB}_{1} $       & -12.6610 & 2.6910 \\
		$ \text{HB}_{2} $       & -12.6610 & 2.6910 \\ \hline
	\end{tabular}}
	\end{center}
\end{table}

\vspace{-0.7cm}
\begin{table}[H]
    \centering \small 
    \captionsetup{font=scriptsize}
	\caption{The average docking results of the SA, GA, LS and HB algorithms after 10 runs for the second case study.}
	\vspace{-0.2cm}
	\begin{center}	
		\scalebox{0.6}{
	\begin{tabular}{|l|l|l|}
		\hline
		Algorithm \hspace{1cm} & Energy \hspace{1cm}  & RMSD \hspace{1cm}  \\ \hline
		Tuned $ \text{SA}^{*}_{1} $      & -15.0124 & 6.2430 \\
		Tuned $ \text{SA}^{*}_{2} $      & -13.0790 & 6.0120 \\
		$ \text{SA}_{1} $       & -14.0790 & 6.5380 \\
		$ \text{SA}_{2} $       & -12.9610 & 6.4640 \\ \hline
		Tuned $ \text{GA}^{*}_{1} $      & -12.728  & 6.722  \\
		Tuned $ \text{GA}^{*}_{2} $      & -12.728  & 6.722  \\
	    $ \text{GA}_{1} $       & -14.0030 & 6.6350 \\
		$ \text{GA}_{2} $       & -12.1460 & 6.4200 \\ \hline
		Tuned $ \text{LS}^{*}_{1} $     & -15.6910 & 6.6340 \\
		Tuned $ \text{LS}^{*}_{2} $     & -13.4850 & 6.4690 \\
		$ \text{LS}_{1} $       & -14.8500 & 6.6480 \\
		$ \text{LS}_{2} $       & 0.1520   & 6.3570 \\ \hline
		Tuned $ \text{HB}^{*}_{1} $      & -16.9080 & 6.6600 \\
		Tuned $ \text{HB}^{*}_{2} $      & -15.6030 & 6.4730 \\
		$ \text{HB}_{1} $       & -16.0210 & 6.7040 \\
		$ \text{HB}_{2} $       & -15.8590 & 6.6160 \\ \hline
	\end{tabular}}
	\end{center}
\end{table}

\section{Conclusion}

A typical docking scenario in Autodock includes applying optimization algorithms. Following those steps, users should select a set of appropriate hyperparameters to maximize the performance of their final results. As it is often beyond the abilities of novice users, we proposed to develop a multi-objective approach for automatically tuning highly parametric algorithms of the Autodock. Automating end-to-end process of applying the proposed MO-SRS offers the advantages of more accurate solutions.  The experimental results clearly show that the obtained results by MO-SRS  outperform hand-tuned models. We hope that the introduced MO-SRS helps non-expert users to more effectively apply Autodock to their applications.

\bibliographystyle{IEEEtran}
{\tiny
\bibliography{conference_041818}}

\end{document}